\documentclass[journal, twocolumn]{IEEEtran}
\IEEEoverridecommandlockouts

\usepackage{graphicx}
\usepackage{xcolor}
\usepackage[skip=1pt]{subcaption}
\usepackage{amsmath}
\usepackage{gensymb}
\usepackage{amssymb}
\usepackage{algorithm}
\usepackage{algorithmicx}
\usepackage[noend]{algpseudocode}
\usepackage{booktabs}
\usepackage{titlesec}
\usepackage{listings}
\usepackage{enumitem}
\usepackage{multirow}
\usepackage{placeins}
\usepackage{rotating}
\usepackage{pifont}
\usepackage{array}
\usepackage{mdframed}
\usepackage{lipsum}
\usepackage[utf8]{inputenc}
\usepackage[many]{tcolorbox}
\usepackage{varwidth}
\usepackage{colortbl}
\usepackage{array}
\usepackage[colorlinks=true, allcolors=blue]{hyperref}
\usepackage{cleveref}
\usepackage{makecell}
\usepackage{duckuments}
\usepackage{comment}

\usepackage[font=scriptsize]{caption}
\usepackage{capt-of,etoolbox}

\makeatletter
\patchcmd\@maketitle\null{{\myfigure{}\par}}{}{}
\makeatother

\title{
\LARGE
DIO: Dataset of 3D Mesh Models of Indoor Objects\\ for Robotics and
Computer Vision Applications
}
\author{
Nillan Nimal$^{\dagger}$,  
Wenbin Li$^{\bullet}$,  
Ronald Clark$^{\ddagger}$,  
Sajad Saeedi$^{\star}$
\vspace{-7 mm}
\thanks{\hspace{-2.4 mm}$^{\dagger}$University of Toronto, corresponding author: nnimal@torontomu.ca, $^{\bullet}$University of Bath, $^{\ddagger}$University of Oxford, $^{\star}$Toronto Metropolitan University
}
}

\begin{document}

\maketitle


\begin{figure*}[th!]
\centering
\includegraphics[width=\linewidth]{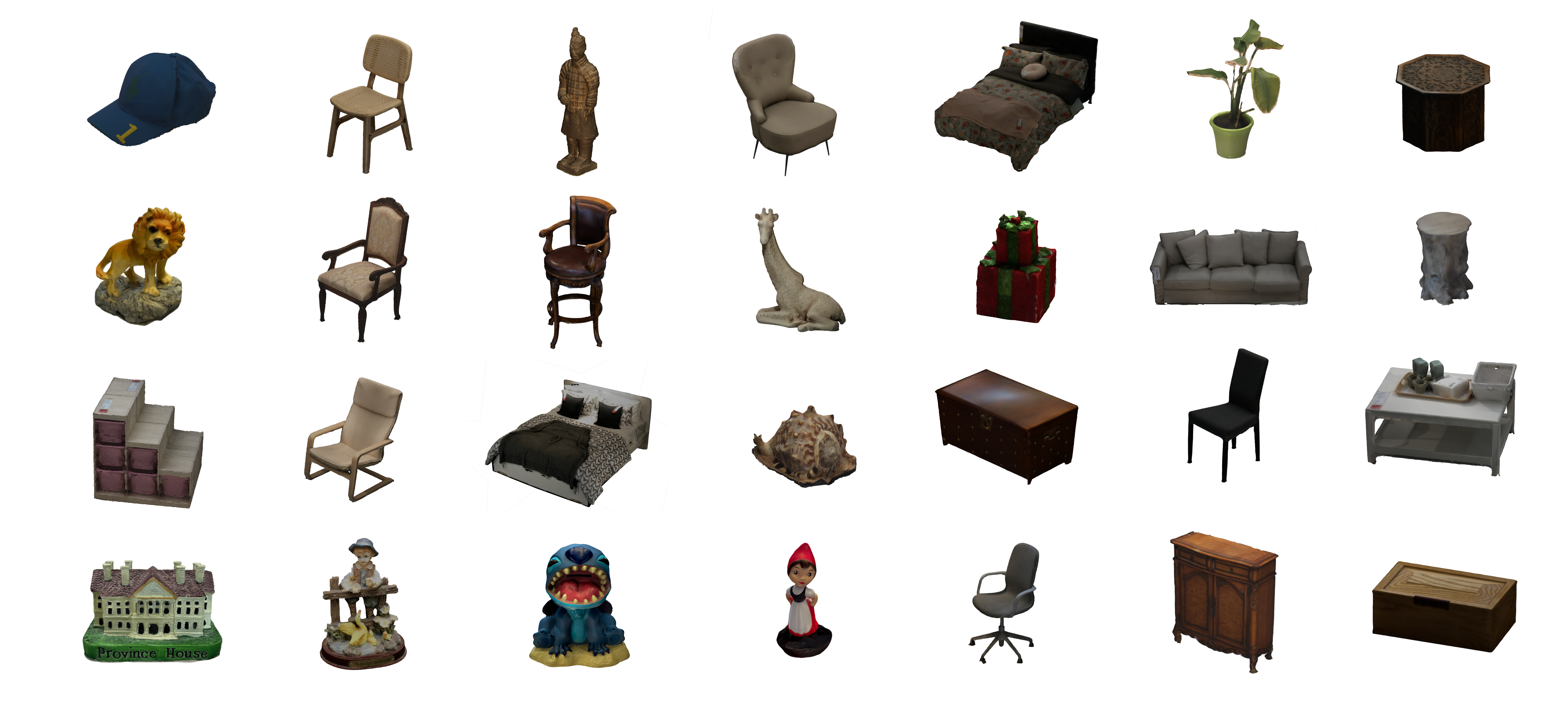}
\caption{Sample meshes available in the dataset.}
\label{fig:sample}
\end{figure*}



\begin{abstract}

The creation of accurate virtual models of real-world objects is imperative to robotic simulations and applications such as computer vision, artificial intelligence, and machine learning. This paper documents the different methods employed for generating a database of mesh models of real-world objects. These methods address the tedious and time-intensive process of manually generating the models using CAD software. Essentially, DSLR/phone cameras were employed to acquire images of target objects. These images were processed using a photogrammetry software known as Meshroom to generate a dense surface reconstruction of the scene. The result produced by Meshroom was edited and simplified using MeshLab, a mesh-editing software to produce the final model. Based on the obtained models, this process was effective in modelling the geometry and texture of real-world objects with high fidelity. An active 3D scanner was also utilized to accelerate the process for large objects. All generated models and captured images are made available on the website of the project.

\end{abstract}




\section{Introduction} \label{sec:introduction}

In robotics and computer vision research, conducting simulations is a critical tool that allows for the identification of algorithmic and technical problems and the verification of proper robot functioning. To this effect, certain applications require simulated environments composed of models of real-world objects such as furniture, stationery, appliances, and office equipment~\cite{Armeni2017arxiv}. \textcolor{black}{The utility of these models is demonstrated by the Stanford University iGibson project~\cite{li2021igibson}, a photo-realistic simulation environment where researchers can train and evaluate robotic agents for a variety of tasks such as navigation or object manipulation. In iGibson, the objects used to construct the different scenes were obtained from open-source datasets.}
    
As a result, the purpose of this paper is to develop a 3D mesh model database of real-world objects. However, the development of accurate models of these objects, especially in terms of the properties of geometry and texture, poses a major challenge. The classical approach for generating these models entails measuring the geometry of objects and using CAD software for modelling. Nevertheless, this approach is disadvantageous due to the tedium, time, and expense involved. \textcolor{black}{For example, the objects available as part of the ReplicaCAD dataset required 900+ hours of professional 3D artist work~\cite{szot2021habitat}. A similar scale of effort was taken to develop InteriorNet~\cite{InteriorNet18}}
    
\textcolor{black}{In order to address the shortcomings of the manual CAD-based method, this research explored different technologies and algorithms to streamline the process of generating photo-realistic 3D models. The approaches documented in this report include photogrammetry and active 3D scanning.} 
\textcolor{black}{The paper presents a detailed classical pipeline to generate high-quality 
 textured mesh models. 
The models generated are from household items and IKEA products. A similar and recent dataset is the Google Scanned Objects (GSO)~\cite{Downs2022GoogleSO} with objects limited to 50cm in size. Our dataset includes large objects such as desks and tables. Moreover, unlike GSO, we present all images, which can be used for training radiance field methods such as Plenoxels~\cite{yu_and_fridovichkeil2021plenoxels}, Neural Radiance Fields (NeRF)~\cite{mildenhall2021nerf}, and 
the 3D Gaussian Splatting~\cite{kerbl3Dgaussians}. Fig.~\ref{fig:sample} demonstrates sample models, presenting both mesh. Please see the website of the project for more information and the generated models\footnote{\href{https://sites.google.com/view/scanned-object-dataset/home}{https://sites.google.com/view/scanned-object-dataset/home}}.}

\textcolor{black}{The rest of the paper is organized as follows.
Sec.~\ref{sec:related} presents the related works. 
Sec.~\ref{sec:Photogrammetry} and Sec.~\ref{sec:active} describe the data collection process with all details for easy reproduction of the paper.
Sec.~\ref{sec:details} presents the details of data and models.
Sec.~\ref{sec:conc} concludes the paper with future directions.
}

\section{Related Work} \label{sec:related}

\textcolor{black}{Datasets of scenes, e.g. ScanNet++~\cite{yeshwanthliu2023scannetpp}, and objects, e.g. ObjectNet3d~\cite{Xiang2016ObjectNet3DAL} and Google Scanned Objects (GSO)~\cite{Downs2022GoogleSO} have many application in robotics and computer vision, such as object recognition~\cite{BigBIRD}, semantic segmentation~\cite{InteriorNet18}, and object manipulation for robotics~\cite{YCB}. Most of such datasets are 
3D CAD models \cite{3dfuture}, \cite{ABO}, \cite{PhotoShape}, \cite{princeton_shape}, \cite{Tatsuma2012}, \cite{Xiang2014BeyondPA}, \cite{Chang2015ShapeNetAI}, \cite{Mo2018PartNetAL} or 
3D scanned models \cite{TanksTemples}, \cite{SHREC11}, \cite{SHREC15}, \cite{SHREC22}, \cite{Sun2018Pix3DDA}.}

\textcolor{black}{Various techniques exist for generating 3D models of objects, including CAD model conversion as in InteriorNet~\cite{InteriorNet18}, using scanning tools as in the digital Michelangelo project~\cite{michelangelo}, and manual modeling~\cite{szot2021habitat} as in Habitat 2.0. Recent radiance fields methods, including non-neural algorithms such as Plenoxels~\cite{yu_and_fridovichkeil2021plenoxels}, and neural algorithms, such as Neural Radiance Fields (NeRF)~\cite{mildenhall2021nerf} are also emerging as fast and reliable modeling methods. Some manufacturers, such as IKEA~\cite{ikea} and Siemens~\cite{siemens} provide CAD models of their products, to aid designers with interior and industrial design, which can also be useful for robotics and computer vision applications.}

\textcolor{black}{Among the existing datasets such as Objaverse~\cite{objaverse}, Google Scanned Objects (GSO)~\cite{Downs2022GoogleSO}, KIT~\cite{KIT}, BigBIRD~\cite{BigBIRD}, and YCB~\cite{YCB}, the dataset presented in this work is closely related to GSO~\cite{Downs2022GoogleSO}. However, GSO and other datasets such as BigBIRD and YCB use scanning rigs that can only accommodate small objects. Notably, GSO does not accommodate objects larger than 50cm. 
Our dataset has a mix of small and large items such as toys, desks, and chairs. Also, GSO does not provide images for their objects, which can be useful for mapping applications, only mesh models are provided. 
Similarly, Objaverse does not provide images of the models.
We present over 3000 high-resolution images that can be used for object detection/classification applications and to generate NeRF models. Our data set presents 141 models in total, 86 are generated from the structure sensor and 55 are photogrammetry models. The dataset has 13 categories, whereas the furniture category has 15 subcategories, such as chairs, desks, and tables.}

\section{Photogrammetry Dataset Generation Process}\label{sec:Photogrammetry}
\textcolor{black}{This section details the initial technique used for dataset creation: photogrammetry, a passive 3D scanning process. Photogrammetry entails capturing a series of overlapping photographs from multiple viewpoints. These images are then processed using photogrammetry software to recover 3D geometry and construct detailed 3D models with textures. This discussion covers various aspects of the methodology, including the configuration of camera setups, the systematic image collection process, model generation, and the steps for post-processing to achieve the final product. }

\subsubsection{Image Set Acquisition } \label{subsec:another}
	
Upon identifying an object to be scanned, the first step in the process involved preparing the object environment. Since images were captured in indoor spaces, lighting was adjusted to avoid producing high-contrast shadows. Essentially, shadows pose issues for the software as there are a limited number of idiosyncratic features that can be matched across images \cite{shadows}. Diffuse lighting was set up in the target object environment through the use of filtered LED lamps on tripods. 

Photogrammetry is highly dependent on the quality and resolution of the images captured. As a result, in terms of equipment, a Canon EOS 5D Mark II DSLR -digital single-lens reflex camera- and a BLACK's BX-75 tripod were employed for image acquisition. The camera was set to manual mode which allows for the ISO, aperture, and shutter speed to be controlled according to the lighting environment to produce the sharpest image. ISO controls the light sensitivity of the camera \cite{camera}. For photogrammetry, a low ISO setting is optimal: image noise is reduced by reducing light sensitivity~\cite{UVL}. The aperture describes the size of the opening of the lens diaphragm which allows light to enter the camera. \textcolor{black}{The aperture size is measured by the f-stop number: the ratio of the focal length to the diameter of the aperture. Consequently, the f-stop number is inversely related to the aperture size; as the f-stop increases, the aperture size decreases. The aperture setting is critical as it is used to control the Depth of Field (the region that is in sharp focus within an image). The Depth of Field increases with a smaller aperture size or larger f-stop. However, at high f-stop settings, image sharpness is lost due to diffraction. Therefore, for photogrammetry, the f-stop must be controlled to ensure that the target object is in sharp focus without approaching the diffraction limit of the lens\cite{camera}}. Based on recommendations made by Autodesk \cite{autodesk} and the University of Virginia Library Scholar's Lab \cite{UVL}, the ISO and f-stop were set to 160 and 8 respectively.  Lastly, the optical zoom for the camera lens was set to the minimum position and was kept constant for all images captured of the object. 

Subsequently, a photo of the X-rite ColorChecker Passport is taken with the camera. The color checker allows for the color profile of the images to be corrected during pre-processing. The ColorChecker Passport also has a scale that is useful for measuring the size of the final model. The color checker is placed next to the object and is captured as part of the image set. 

\begin{figure}[t!]
\centering
\includegraphics[width=\linewidth]{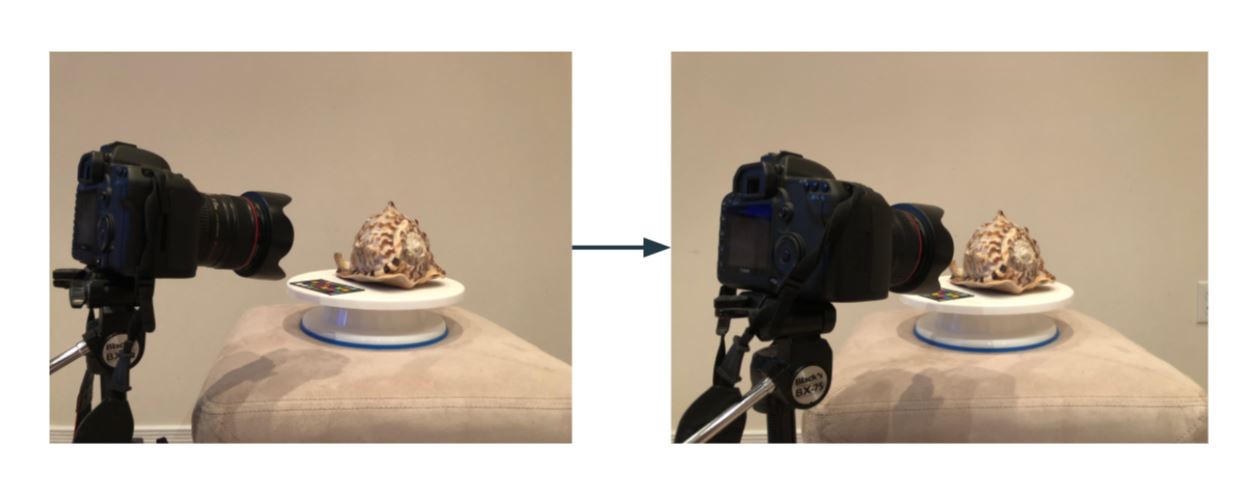}
\caption{The process of moving the camera and tripod globally around the target object for image acquisition.}
\label{fig:setup}
\vspace{5 mm}
\centering
\includegraphics[width=\linewidth]{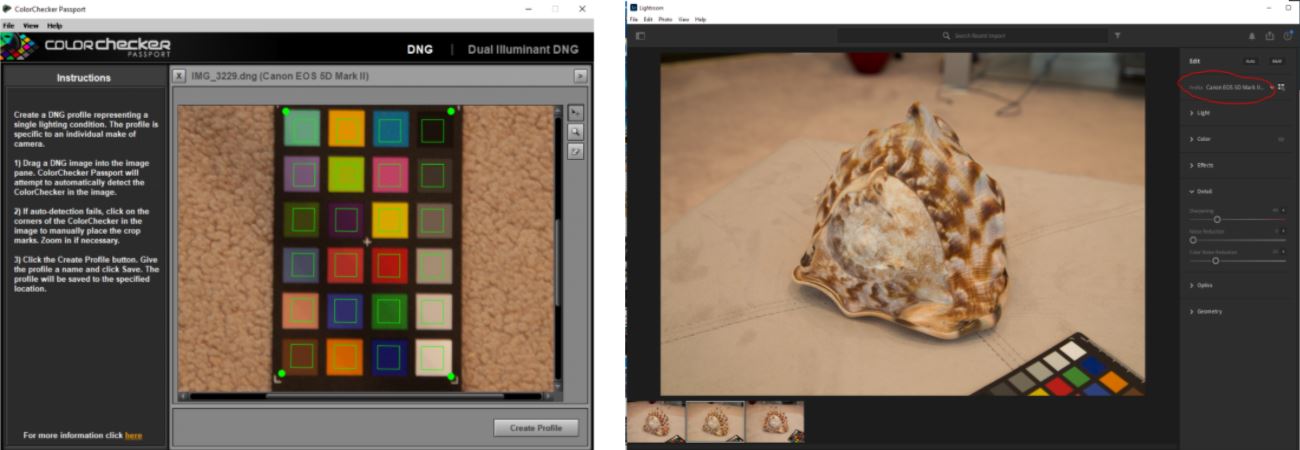}
\caption{Generating the corrected colour profile using the X-rite Calibration software (left) and applying the corrected colour profile in Adobe Lightroom (right). }
\label{fig:color_correction}
\vspace{5 mm}
    \centering
    \begin{subfigure}[b]{0.33\linewidth}
        \includegraphics[width=\linewidth]{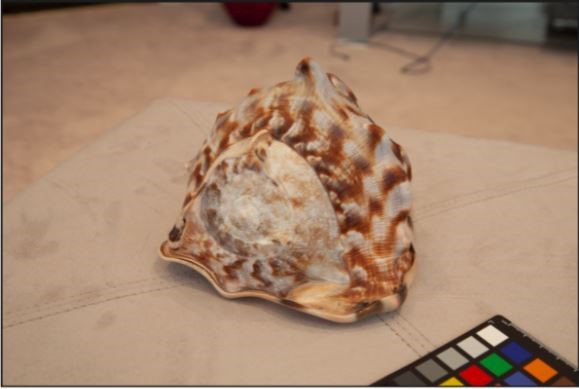}
        \caption{Original Image}
        \label{fig:original_image}
    \end{subfigure}%
    \begin{subfigure}[b]{0.33\linewidth}
        \includegraphics[width=\linewidth]{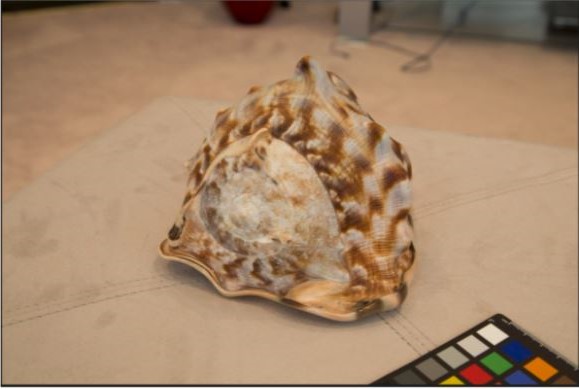}
        \caption{Corrected Image}
        \label{fig:corrected}
    \end{subfigure}%
    \begin{subfigure}[b]{0.33\linewidth}
        \includegraphics[width=\linewidth]{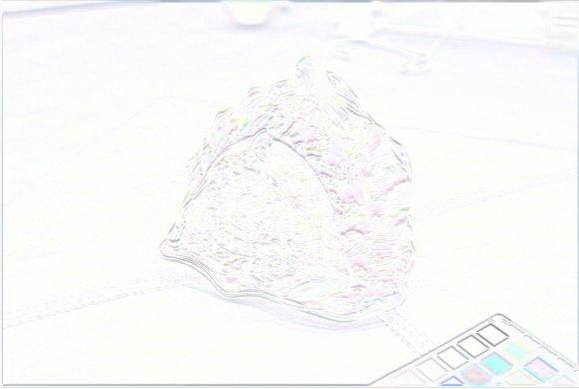}
        \caption{Absolute Difference}
        \label{fig:abs_difference}
    \end{subfigure}
    \caption{The difference between the standard and corrected colour profile for the Canon EOS 5D Mark II.}
    \label{fig:color_correction_after}
\end{figure}

To capture images, the tripod height was initially fixed to align the camera lens with the center of the object. The focus was manually adjusted and the shutter speed was set such that the exposure was neutral according to the light meter of the camera. The focus and shutter speed was adjusted for each picture. After each photo was taken, the tripod was circularly moved around the object as shown in Fig.~\ref{fig:setup}. The degree of overlap between photos impacts the quality of the model reconstruction. As a result, the angular position of the tripod was moved less than 30 degrees between images according to Autodesk recommendations \cite{autodesk}. The position was repeatedly adjusted until a loop was completed. Then the height of the tripod was altered and this process of circling the object was repeated. The number of times the height of the tripod was adjusted and loops were completed was dependent on the size and complexity of the object. 
    
\textcolor{black}{In environments where the tripod and DSLR could not be set up due to space limitations or where professional camera equipment was not permitted (i.e retail stores), a phone camera method was employed. The process of using the ColorChecker Passport and circling the object to obtain images from different viewpoints was similar to the DSLR method discussed previously. However, to compensate for the lower image quality, a higher number of pictures of the target object would be taken compared to using the DSLR. For this method, the 12-megapixel iPhone 8 Plus camera was utilized.}

\subsubsection{Pre-processing: Colour Correction} \label{subsec:yetanother}

In the pre-processing stage, for each object, the picture of the colour checker taken with the camera was imported into the X-rite Camera Calibration software. The software generates a corrected colour profile for the camera used to take the images. This colour profile was imported into Adobe Lightroom and applied to all images as shown in Fig.~\ref{fig:color_correction}. Fig.~\ref{fig:color_correction_after} illustrates the difference between the accurate colour profile and the standard colour reproduction from the Canon DSLR. Consequently, the use of a colour checker is essential to ensure the colour accuracy of the final model. 

Additionally, the images in the set were checked and blurry or out-of-focus images were removed. This is because poor-quality images can severely impact the quality of the model reconstruction. Then, the images, which were originally in the CR2 (RAW) format were exported from Lightroom as TIFF images.

\subsubsection{Meshroom Workflow}
\label{subsec:last}
The pre-processed images are then imported into Meshroom. Referencing a Camera Sensors Database, Meshroom determines the internal parameters of the cameras used to capture the images and groups them accordingly \cite{meshroom}. Images with missing metadata are flagged. These images, in which Meshroom failed to extract metadata and create camera intrinsics, are manually removed from the project. Then the reconstruction is run using the standard Meshroom pipeline.

\subsubsection{Post-processing}
\label{subsec:lastanother}

\begin{figure}[h!]
  \centering
  \subfloat[Raw Mesh]{\includegraphics[width=\linewidth]{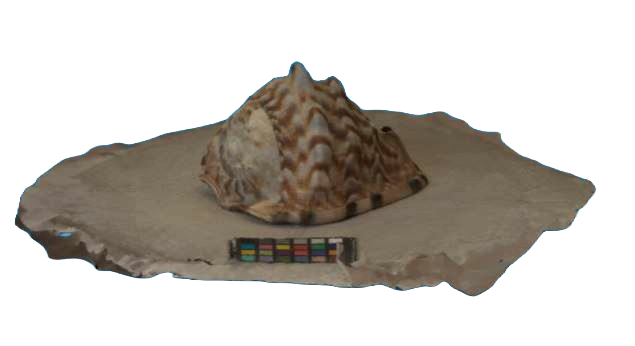}\label{fig:f1}}
  \hfill
  \subfloat[Edited Model]{\includegraphics[width=0.6\linewidth]{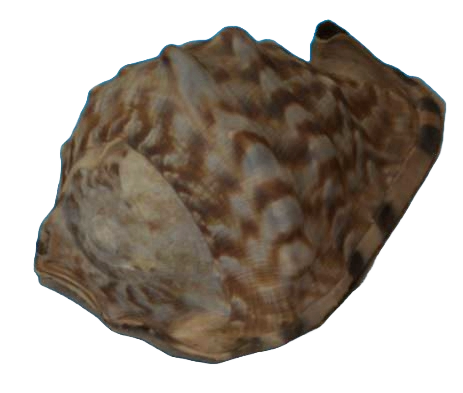}\label{fig:f2}}
  \caption{Editing the raw mesh model of a shell using MeshLab.}
  \label{fig:mesh}
\end{figure}

To process the result produced by Meshroom, MeshLab, an open-source 3D mesh editing software was utilized~\cite{meshlab}. \textcolor{black}{The raw mesh as seen in Fig.~\ref{fig:mesh}-a is scaled to real-world dimensions by measuring the ColorChecker passport scale, which is included in the reconstruction of the target object using the MeshLab Measuring Tool. By comparing the measured distance between the scale increments for the ColorChecker in MeshLab to the real-world distance, the scaling factor can be computed and applied to the mesh. Once the mesh is scaled, the background noise and the reconstructed ColorChecker are removed by using the Area Selection Tool to isolate the target object. The following step is mesh simplification. Initially, the result from Meshroom is a high-density mesh that is high in polygon count. As a result, the Quadric Edge Collapse Decimation feature was used in MeshLab to perform polygon reduction on the mesh while preserving the original geometry. This is advantageous as editing a high-density mesh is computationally taxing for the software and the mesh file is large in size.  MeshLab smoothing filters are then used to smooth rough surfaces on the model. Finally, the mesh is centered with respect to the MeshLab origin and aligned such that the object is in the correct orientation. This final result is shown in Fig.~\ref{fig:mesh}-b}

\section{Active 3D Scanning Dataset Generation Process}\label{sec:active}

\textcolor{black}{To address the time-consuming nature and computational demands of scanning objects via photogrammetry, we also adopted an active 3D scanning approach. This method relies on devices that emit radiation and detect the reflection or transmission of the radiation to probe an object and recover depth data for the reconstruction of geometry. Specifically, this project utilized the Occipital Structure Sensor Mark II Pro, an active 3D scanning device that attaches to an iPad and uses structured-light technology in conjunction with the iPad's RGB camera to scan objects 
This section details the calibration procedure, scanning process, and post-processing pipeline used for the Structure Sensor.}

\subsection{Calibration}
The Structure Sensor was calibrated using the indoor calibration mode in the Occipital Calibrator App. A complex scene was selected that offered high contrast in both color and infrared feeds to ensure accurate calibration 
Then, the calibration was refined by adjusting the alignment between the depth and color feeds. 

\subsection{Scanning Process}
\begin{figure}[h]
    \centering
    \begin{subfigure}{0.3\textwidth}
        \centering
        \includegraphics[width=\linewidth]{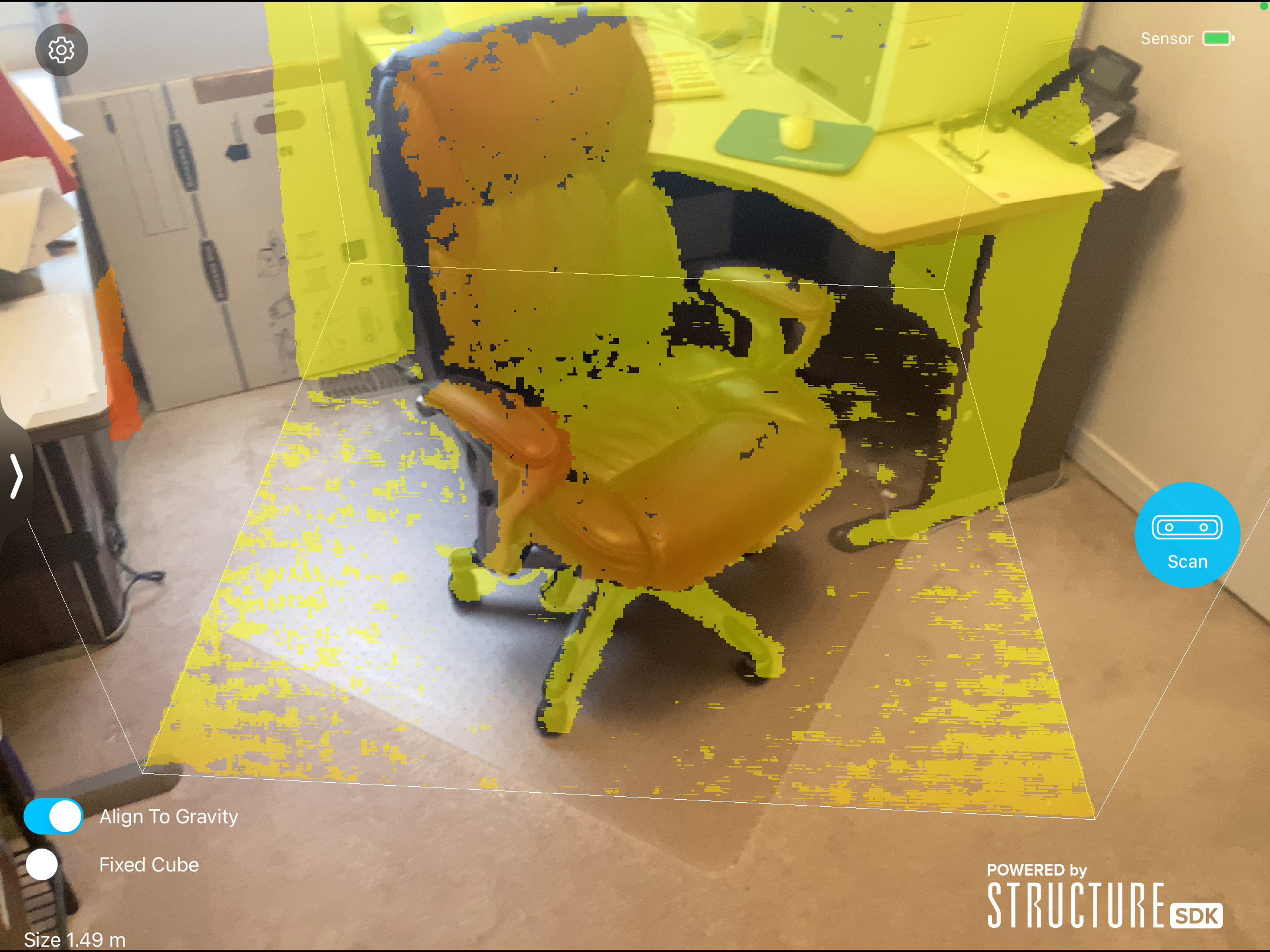} 
        \caption{Setting the bounding box.}
        \label{fig:bbox}
    \end{subfigure}
    \hfill
    \begin{subfigure}{0.3\textwidth}
        \centering
        \includegraphics[width=\linewidth]{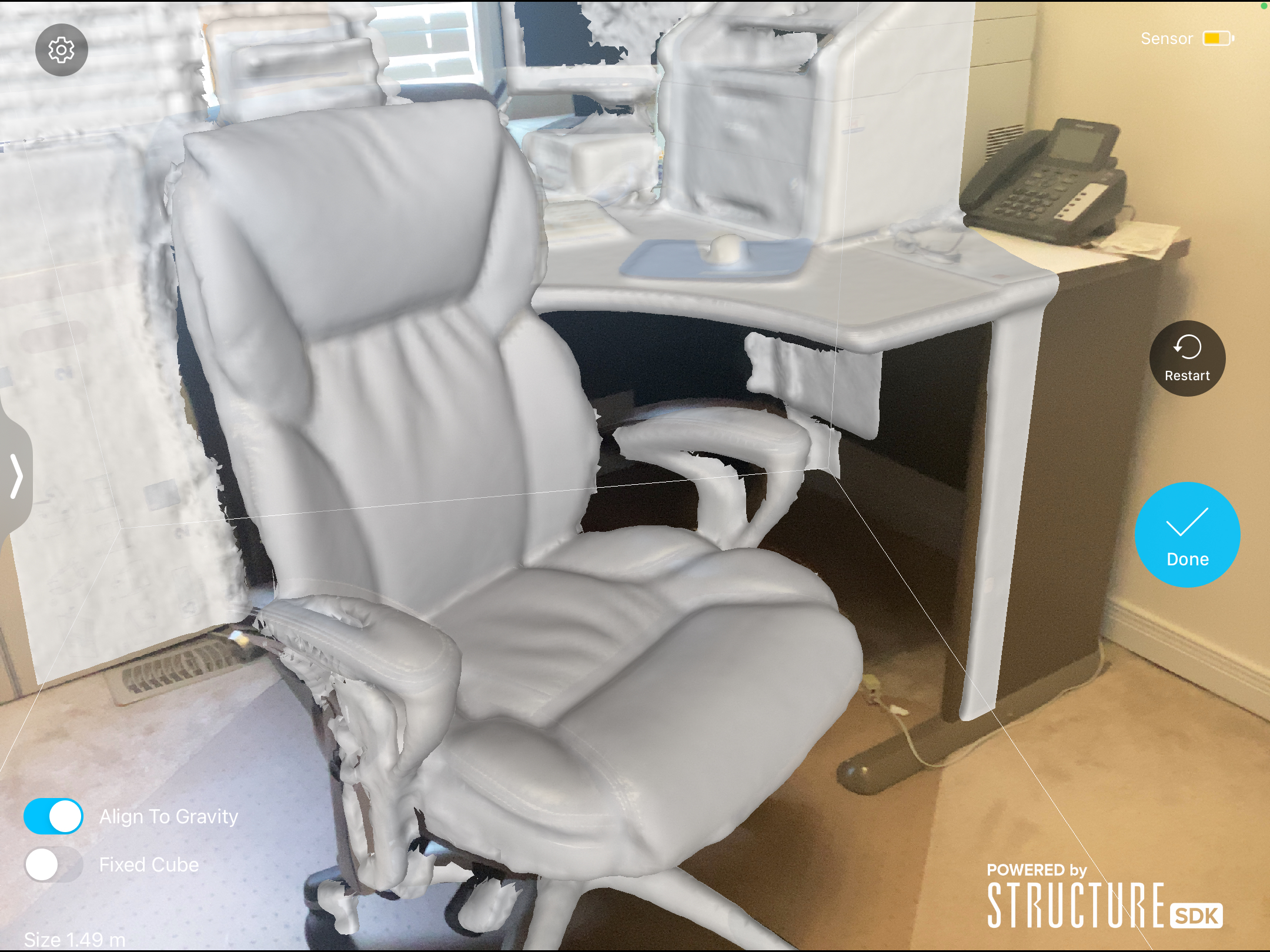} 
        \caption{Scanning the object.}
        \label{fig:scanning}
    \end{subfigure}
    \hfill
    \begin{subfigure}{0.3\textwidth}
        \centering
        \includegraphics[width=\linewidth]{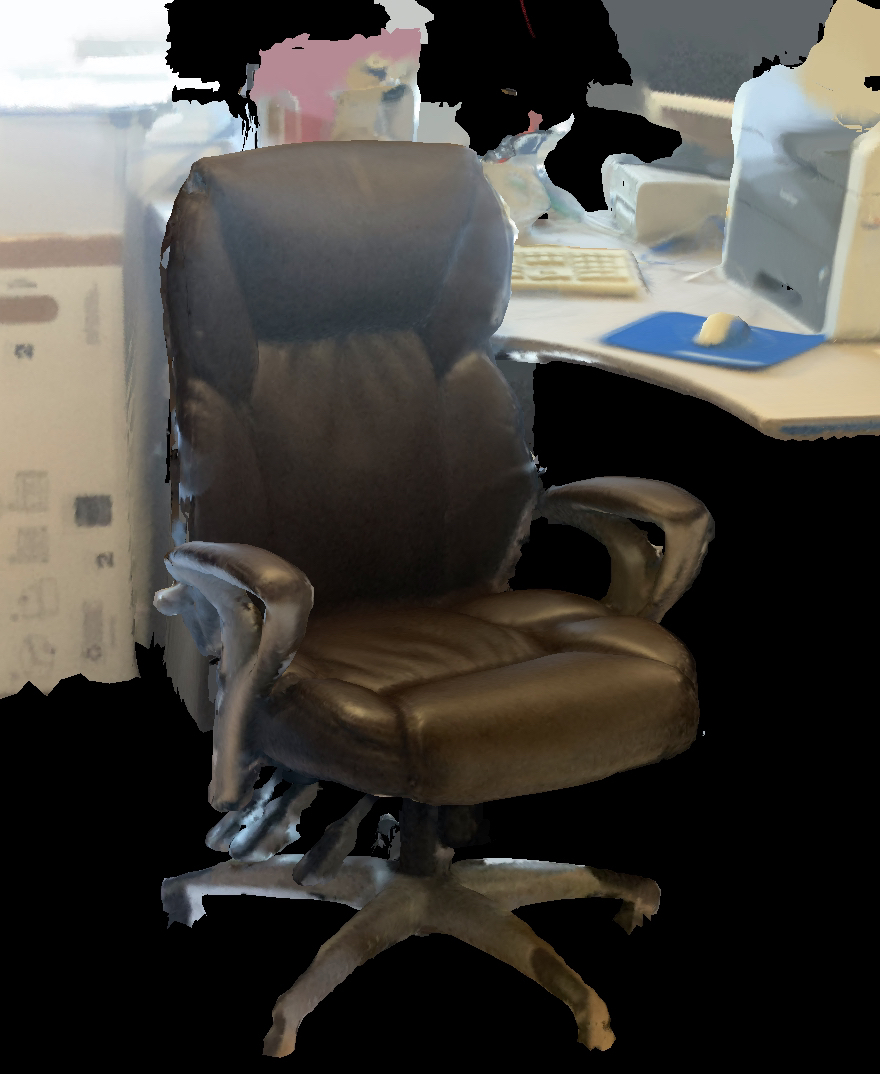} 
        \caption{Raw output from the Structure Sensor application.}
        \label{fig:output}
    \end{subfigure}
    \caption{The Structure Sensor scanning process for an office chair. }
    \label{fig:Scanning Process}
\end{figure}

The scanning process begins by setting a bounding box that encapsulates the target object (Fig.~\ref{fig:bbox}). Then the iPad with the attached Structure Sensor is moved around the object slowly. During this process, images are captured of the target object for texture generation. Parts of the object that have been captured by the Structure Sensor are coloured grey in the iPad color feed. This real-time feedback allows for the identification of missed areas in the scan, preventing large holes and geometric inaccuracies that would need to be corrected in post-processing. This is one advantage of using the Structure Sensor over photogrammetry, since missed parts of the object would only be known after processing the images with the photogrammetry software. Once the entire model is covered and the scanning process is complete, the captured data is processed to generate a textured model using the iPad hardware. 

\subsection{Post-processing}
The post-processing for the Structure Sensor models is similar to the process mentioned previously for the photogrammetry models, with a few exceptions. The output from the Structure Sensor is already a simplified mesh, scaled to real-world dimensions. As a result, the steps of decimating and scaling the mesh in MeshLab are unnecessary.  

\section{Features of the Dataset} \label{sec:details}
This section describes the key features and properties of the dataset.
\subsection{Statistics}
\begin{figure}[t!]
\centering
\includegraphics[width=\linewidth]{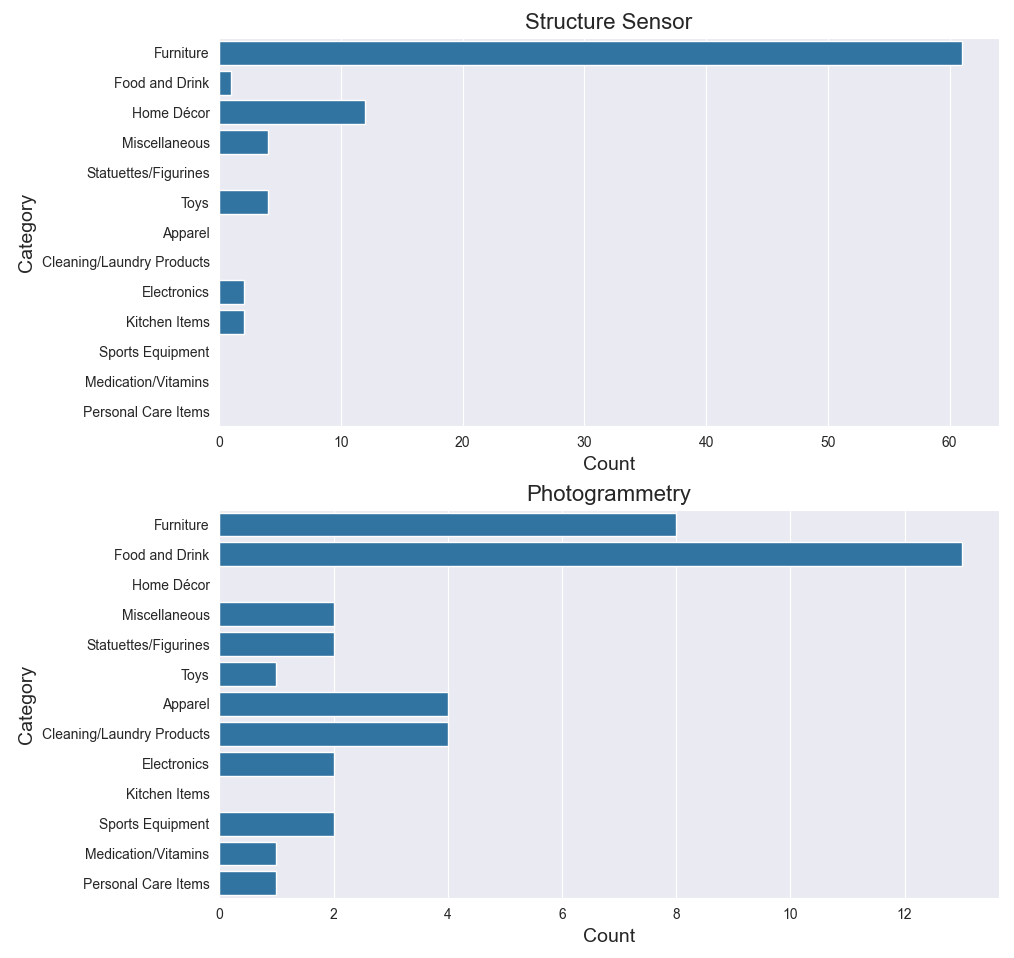}
\caption{Plots demonstrating the distribution of models across categories for the Structure Sensor and Photogrammetry Pipelines.}
\label{fig:categories_plot}
\end{figure}

The dataset is categorized into 13 categories as shown in Table \ref{table:objects_category}. As shown in Fig.~\ref{fig:categories_plot}, a predominant portion of the Structure Sensor objects are in the Furniture and Home Décor categories, which are mostly large items. In contrast, the highest number of photogrammetry models are in the Food and Drink category, with other models distributed over the categories of smaller objects such as Apparel, Toys, Electronics, etc. This distribution pattern is due to the fact that the Structure Sensor made scanning larger objects (i.e., furniture items) easier where it would be difficult to ensure proper coverage of the object using the photogrammetry method. Instead, the photogrammetry method was useful for applications with smaller objects where there was space to set up the DSLR and tripod and orbit around the object. The photogrammetry method could also capture smaller objects (small figures) that could not be scanned with the Sensor Sensor due to its limited scan resolution. 

\begin{figure}[t!]
\centering
\includegraphics[width=\linewidth]{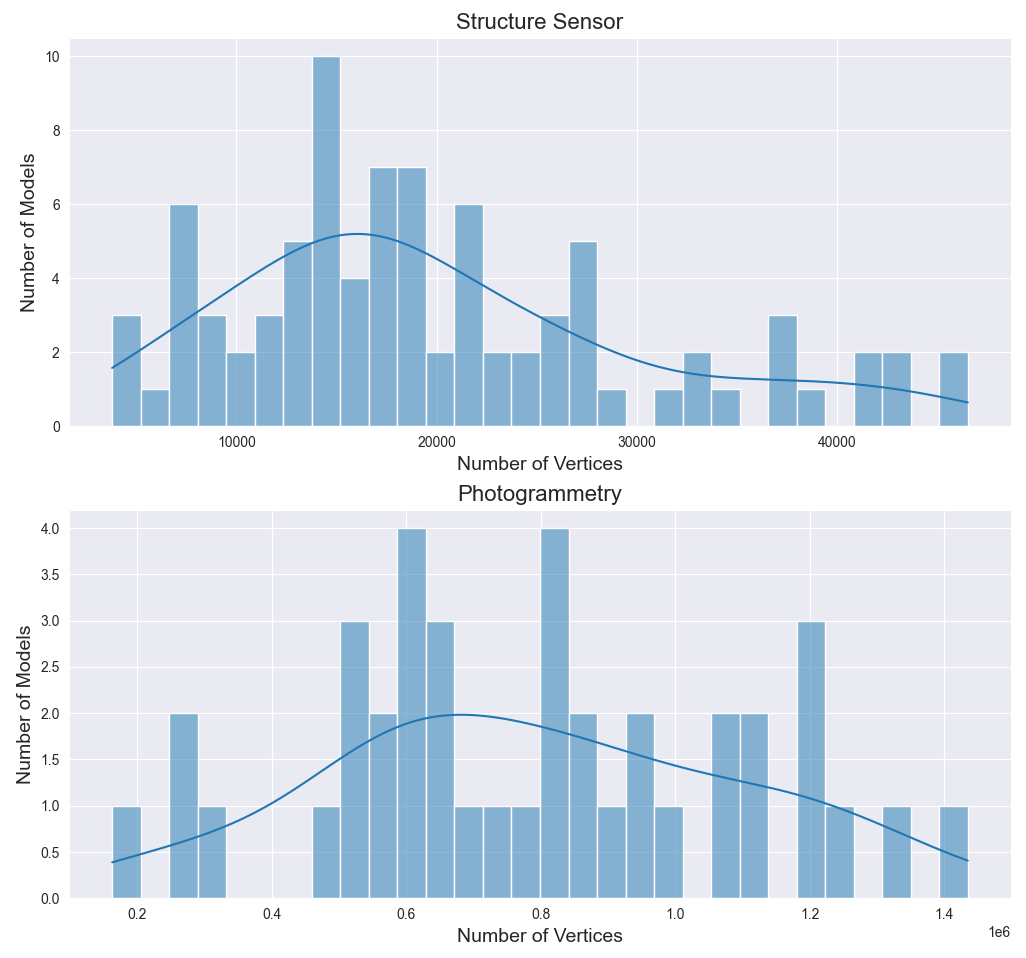}
\caption{Histograms showcasing the distribution of model complexity for both scanning pipelines.}
\label{fig:histograms}
\end{figure}

Model complexity in the dataset was assessed by measuring vertex count. The Structure Sensor models had a lower complexity with a mean vertex count of \(20012 \pm  10326\) (min: 3801, max: 46530)  vs. \(701753 \pm 383371\) (min: 21180, max: 1434864) for the photogrammetry pipeline (see Fig.~\ref{fig:histograms}). A similar trend is seen in terms of the .OBJ file sizes where the Structure Sensor models have an average size of \(2.8 \pm 1.2\) MB (min: 0.5, max: 5.6) vs. \(101.1 \pm 56.3\) MB (min:4.7, max: 203.7) for the photogrammetry pipeline. Although the complexity and sizes of the photogrammetry models are higher than the structure sensor, the photogrammetry models provide higher resolution textures due to the higher resolutions of the DSLR/iPhone~8 Plus cameras compared to the iPad camera working with the Structure Sensor. Note that the Structure Sensor does not output any images.

\begin{table}[h]
\centering
\caption{Number of Objects in Each Category}
\begin{tabular}{|l|c|}
\hline
\textbf{Category}                    & \textbf{Number} \\ \hline
Apparel                              & 4               \\ \hline
Cleaning/Laundry Products            & 4               \\ \hline
Electronics                          & 4               \\ \hline
Food and Drink                       & 16              \\ \hline
Furniture                            & 74              \\ \hline
Home Décor                           & 12              \\ \hline
Kitchen Items                        & 2               \\ \hline
Medication/Vitamins                  & 1               \\ \hline
Miscellaneous                        & 9               \\ \hline
Personal Care Items                  & 1               \\ \hline
Sports Equipment                     & 2               \\ \hline
Statuettes/Figurines                 & 7               \\ \hline
Toys                                 & 5               \\ \hline
\textbf{Total}                       & 141             \\ \hline
\end{tabular}
\label{table:objects_category}
\end{table}

\subsection{Dataset Strengths}
Our dataset includes a wide variety of objects of various sizes, from smaller figurines (5cm wide) to large sectional sofas (200cm wide). Additionally, our pipelines were able to capture intricate geometric details and complex textures to produce high-quality models. We also provide 3584 high-resolution images of objects captured using the photogrammetry pipeline, which is useful for object detection/classification and neural rendering applications. Lastly, as mentioned in \cite{Downs2022GoogleSO}, since the objects were 3D scanned, they are free from idealizations and contain surface imperfections which make them more realistic than hand-modeled objects. 

\begin{figure}[h]
    \centering
    \begin{subfigure}{0.3\textwidth}
        \centering
        \includegraphics[width=\linewidth]{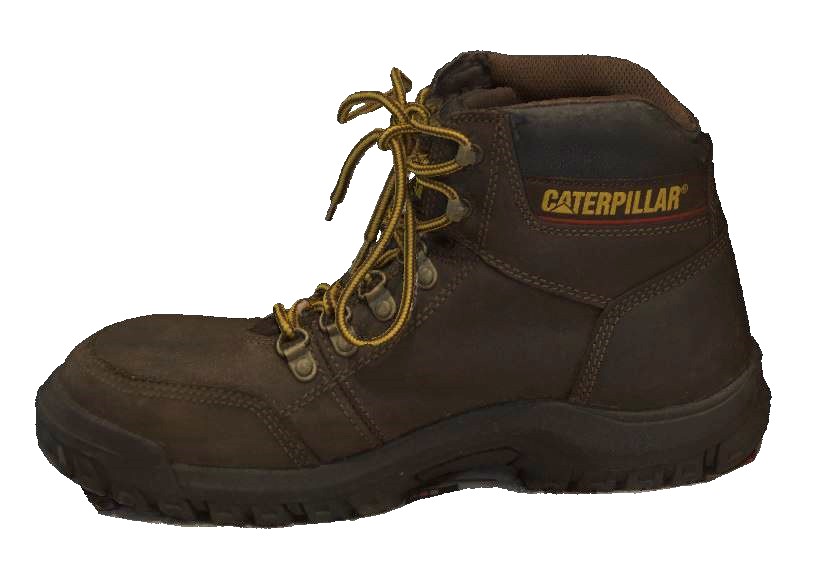} 
        \caption{}
        \label{fig:subfigure1}
    \end{subfigure}
    \begin{subfigure}{0.45\textwidth}
        \centering
        \includegraphics[width=\linewidth]{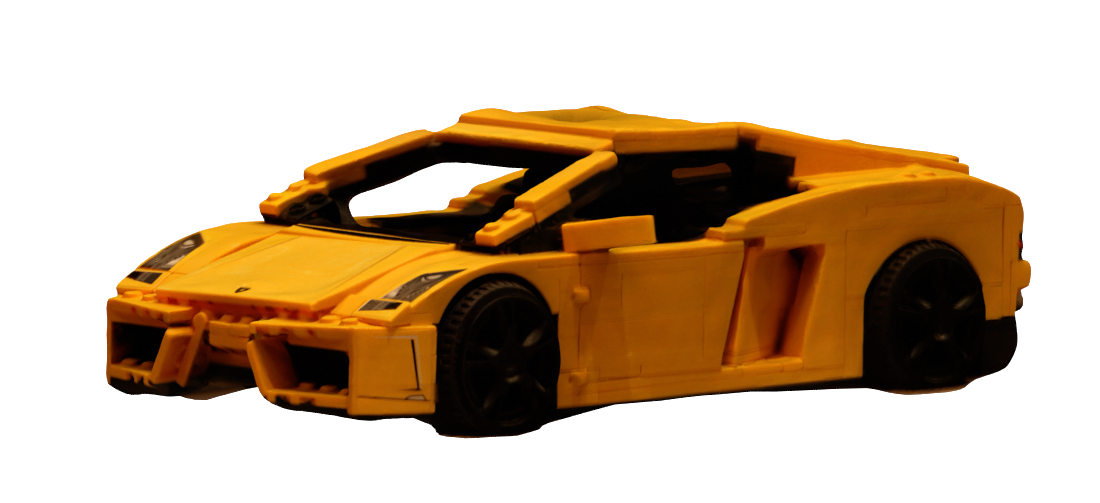} 
        \caption{}
        \label{fig:subfigure2}
    \end{subfigure}
    \vspace{1em} 
    \begin{subfigure}{0.45\textwidth}
        \centering
        \includegraphics[width=\linewidth]{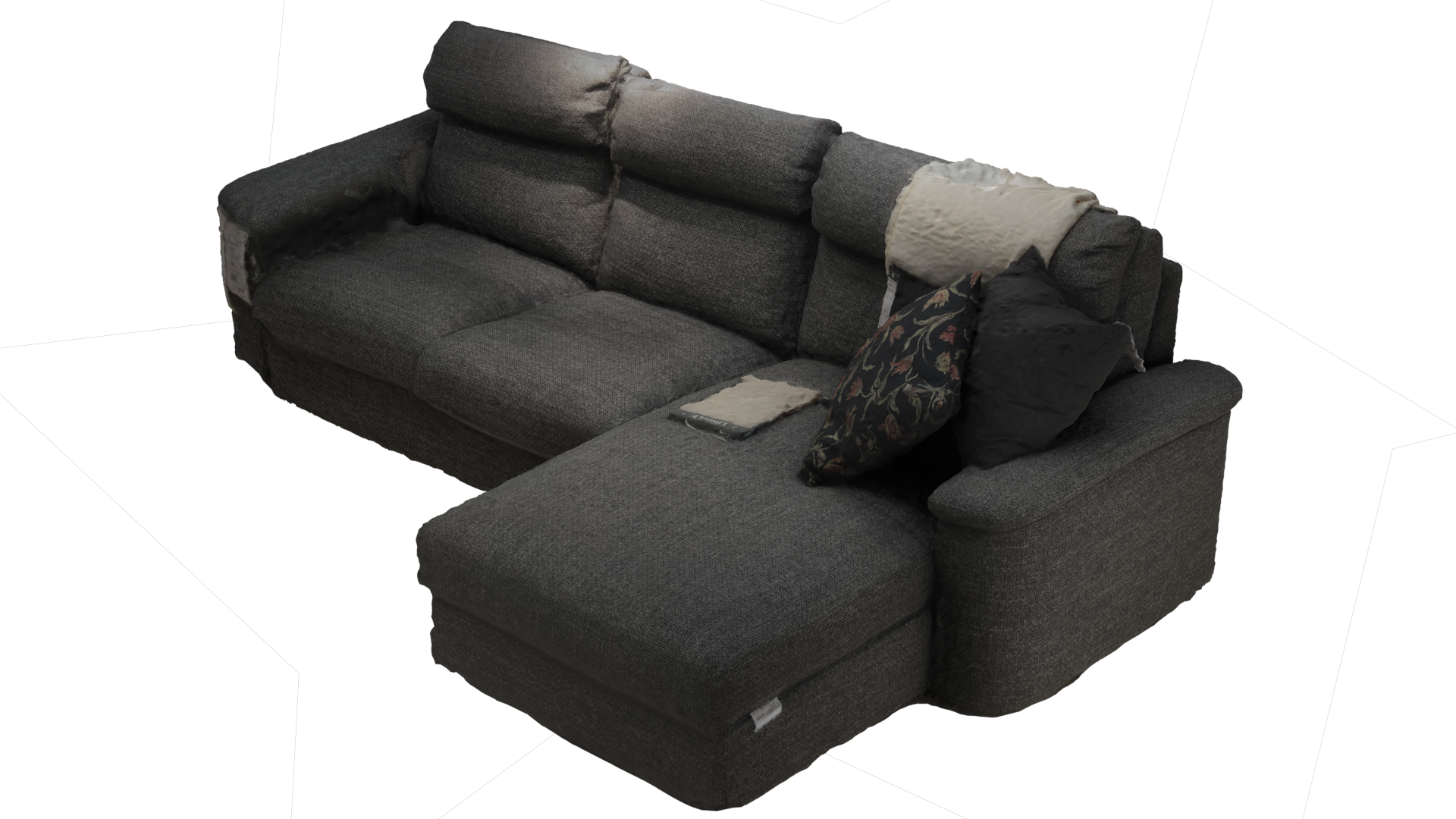} 
        \caption{}
        \label{fig:subfigure3}
    \end{subfigure}
    \begin{subfigure}{0.45\textwidth}
        \centering
        \includegraphics[width=\linewidth]{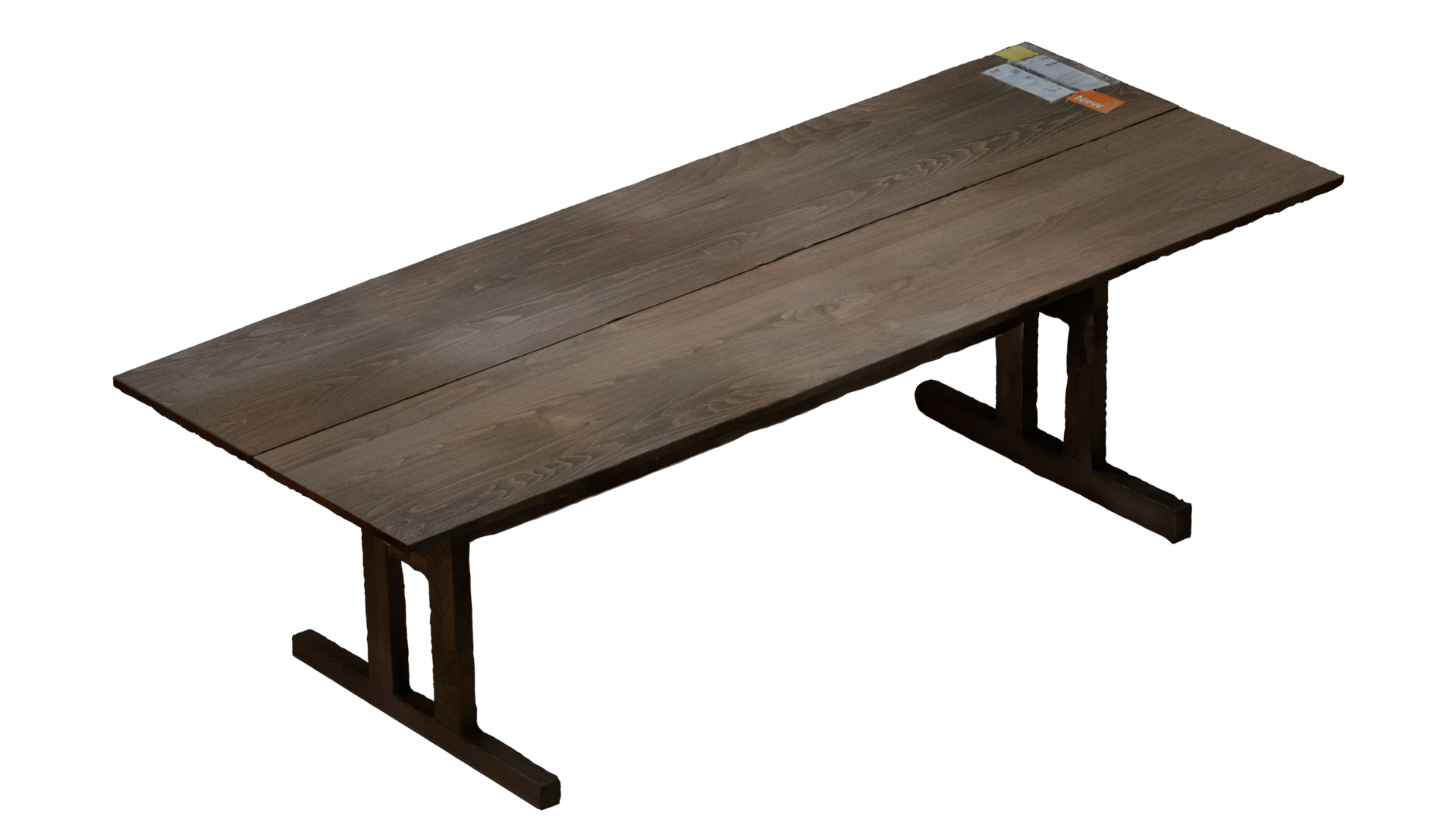} 
        \caption{}
        \label{fig:subfigure4}
    \end{subfigure}
    \caption{Our dataset showcases a diverse range of models, each characterized by intricate geometric details and high-resolution textures. These include: (a) the detailed lacing on a steel-toed shoe, showcasing geometric complexity; (b) the precisely rendered individual bricks of a Lego Lamborghini, highlighting meticulous detail; (c) the realistic fabric texture of a sectional sofa, demonstrating high-resolution texture capabilities; and (d) the authentic wood grain of a dining table, exemplifying our pipeline's ability to capture natural textures.}
    \label{fig:mainfigure}
\end{figure}

\subsection{Dataset Weaknesses}
The pipelines employed in this project were unable to capture objects with reflective, translucent, or transparent surfaces. Scans of these objects were often deformed, incomplete, and inaccurate. Furthermore, compared to other datasets of 3D scanned objects such as GSO \cite{Downs2022GoogleSO} with over 1000 objects, our dataset is limited in size. 

\section{Gazebo Simulations}
\begin{figure}[t!]
\centering

\begin{subfigure}{0.24\textwidth}
\includegraphics[width=\linewidth]{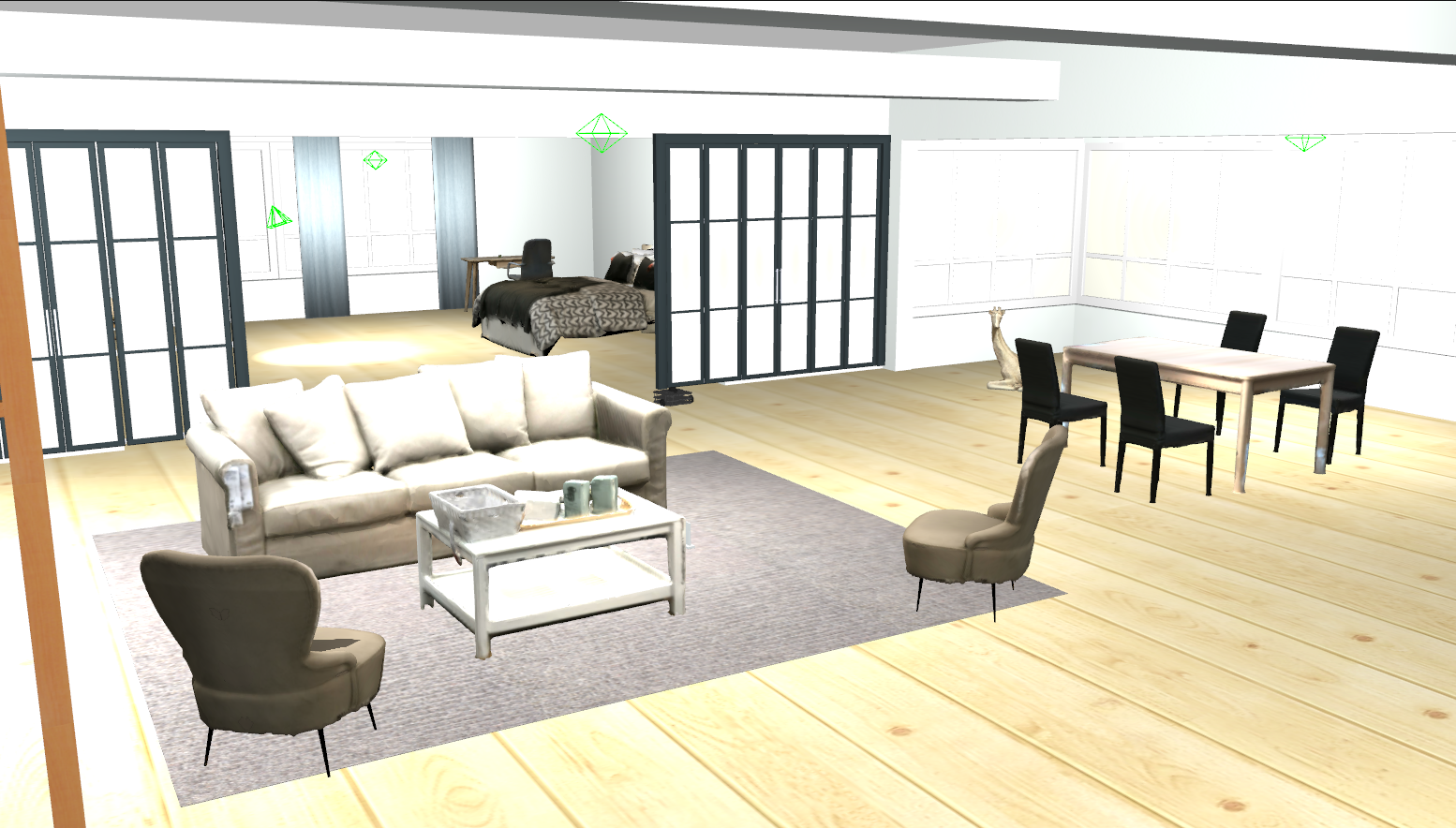}
\end{subfigure}%
\begin{subfigure}{0.24\textwidth}
\includegraphics[width=\linewidth]{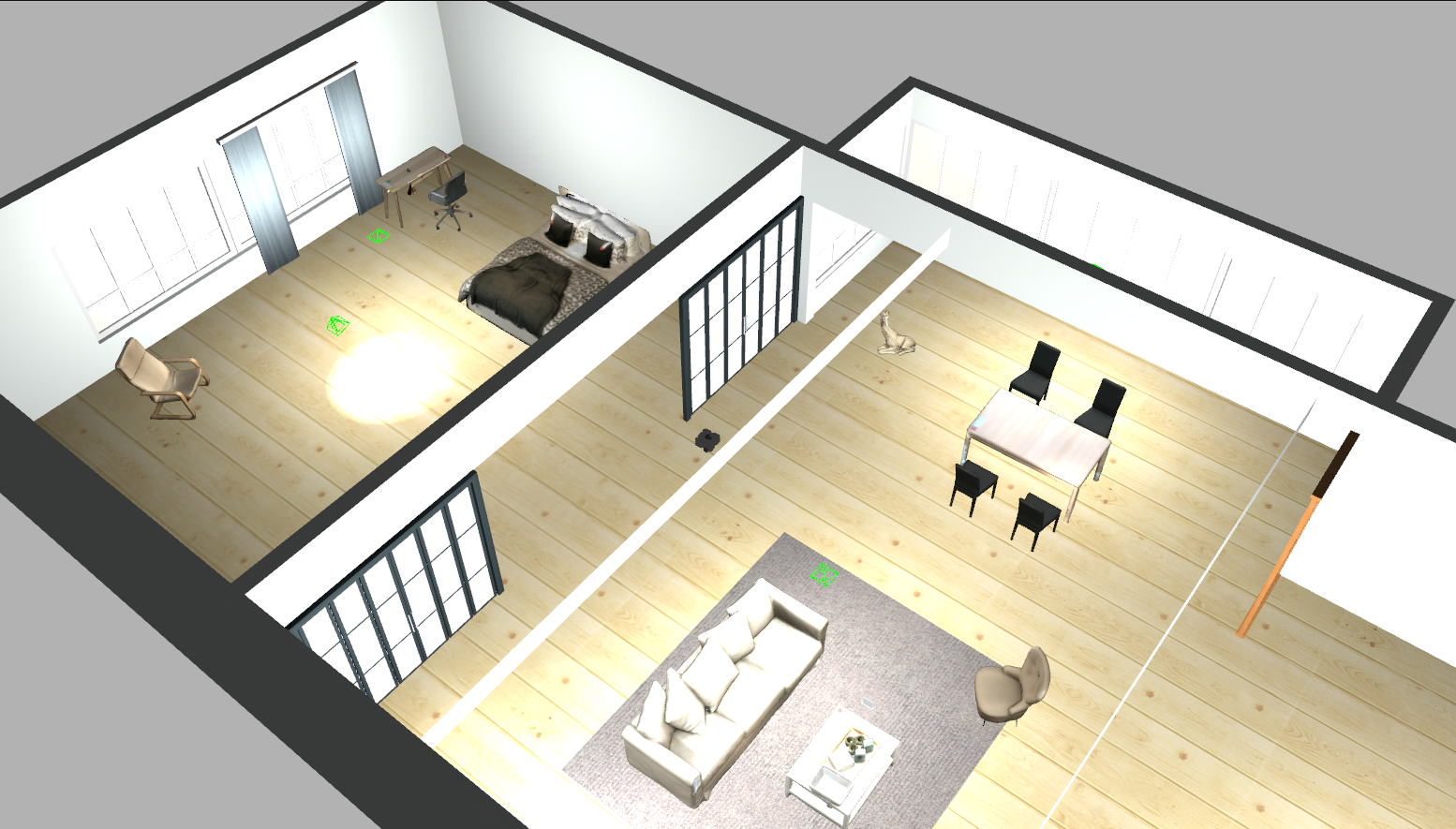}
\end{subfigure}

\begin{subfigure}{0.24\textwidth}
\includegraphics[width=\linewidth]{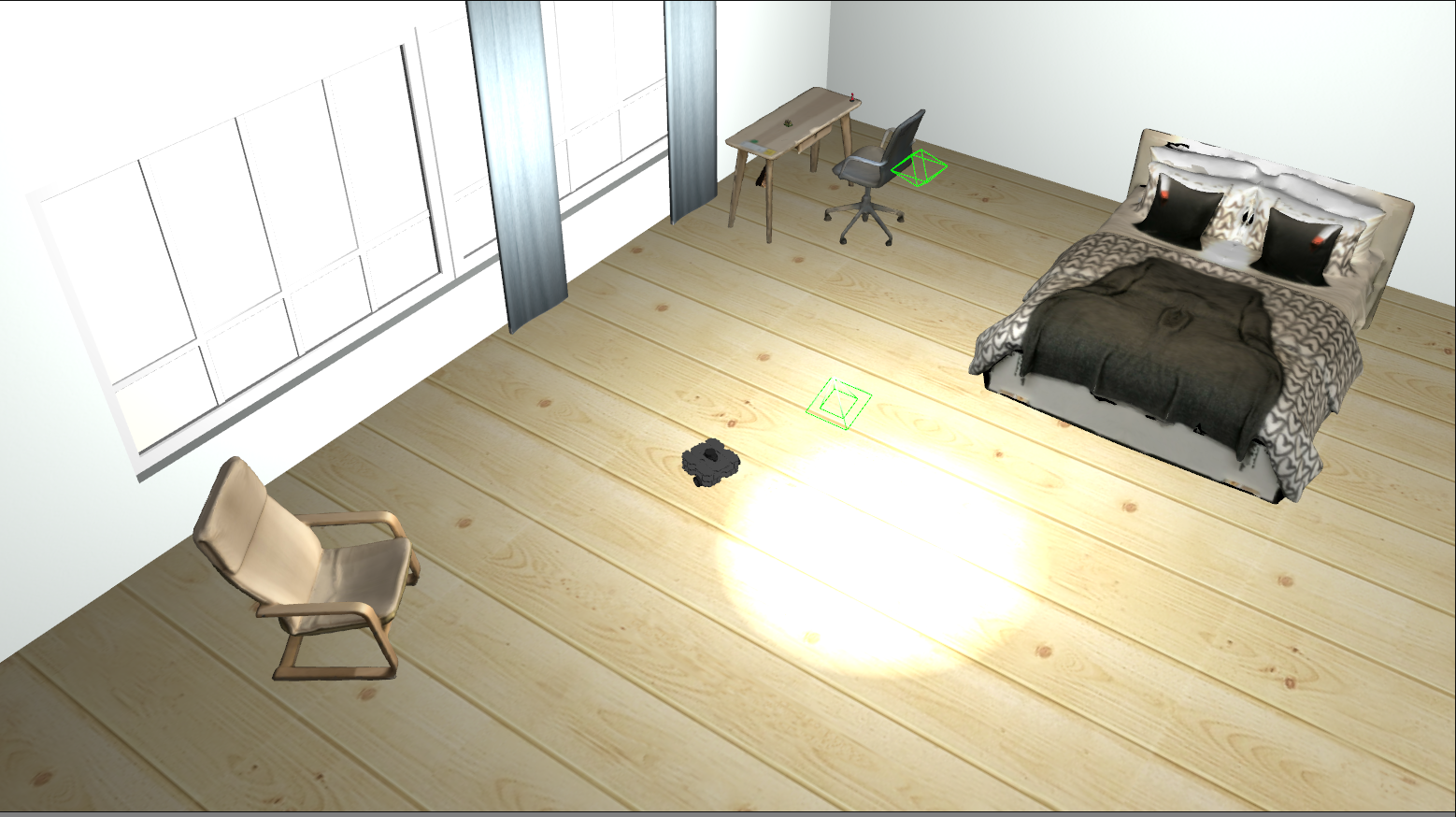}
\end{subfigure}%
\begin{subfigure}{0.24\textwidth}
\includegraphics[width=\linewidth]{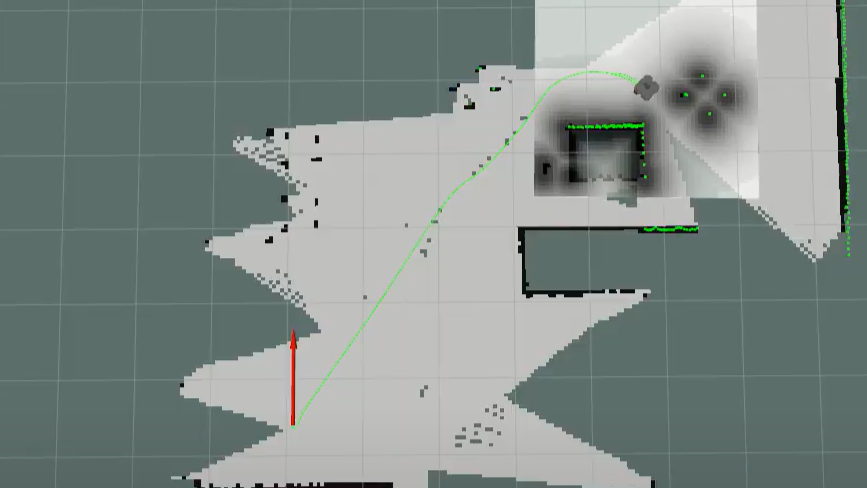}
\end{subfigure}
\vspace{2 mm}
\caption{
A SLAM simulation of a TurtleBot3 Waffle Pi in a customized AWS-Robotics Small House World environment, featuring furniture items from our dataset. All furniture items in the Gazebo world shown are from our dataset. See the video of the simulation on the website of the project: {\href{https://sites.google.com/view/scanned-object-dataset/home}{https://sites.google.com/view/scanned-object-dataset/home}.}
}
\vspace{2 mm}
\label{slam_sim}
\end{figure}

In order to demonstrate the practicality of the models in the dataset, a simulation environment was constructed in Gazebo \cite{koenig2004design}. An AWS-robotics Small House World \cite{aws-robomaker-small-house-world} was utilized as the base, where all the original items were removed and replaced with our future items, as shown in Fig.~\ref{slam_sim}. Only the doors, floor, carpet, and walls were kept. A TurtleBot3 Waffle Pi \cite{turtlebot3_simulation} was deployed to conduct a SLAM simulation in the custom environment, which demonstrates a useful application of our dataset: populating simulation environments with realistic models.

The process of importing the models into the Gazebo world involved converting dataset meshes into SDF (Simulation Description Format) model objects. The SDF models were composed of one link with identical visual and collision elements derived from the dataset's meshes. Large items such as sofas and beds were assumed to be immovable and were set as static. Conversely, inertial properties were added for smaller items in the dataset. 

Inertial properties for the smaller objects were estimated by first ensuring that a given model was watertight. This involved fixing non-manifold geometry and closing holes in MeshLab. Subsequently, the mass of the real-world object was measured and the inertial properties and center of mass of the object were computed using Meshlab based on the assumption that the object is solid and homogeneous. 
These characteristics were integrated into the SDF models, enhancing their utility for applications like pick-and-place operations. Currently, the incorporation of dynamic properties is limited to select items, however, extending it to all smaller objects in the dataset is a prospective direction for future work.

Additionally, to further refine our dataset, we contemplate incorporating simpler collision meshes for each object. This enhancement would optimize the balance between computational efficiency and physical accuracy, especially in dynamic simulations where complex geometries can be computationally intensive.

\section{Conclusions} \label{sec:conc}
This paper presented a dataset of 3D mesh models of indoor objects for robotics and computer vision applications with 141 objects across 13 categories and 3584 images. High-quality textured models were produced using two separate pipelines; photogrammetry with DSLR/iPhone 8 Plus cameras and active 3D scanning with a Structure Sensor Mark II Pro. Each pipeline had unique strengths that allowed for a diverse variety of objects to be captured for the dataset. 

We believe that our models and images are useful for robotics simulation as well as emerging neural rendering methods requiring access to raw imagery data. We plan to continue improving the models in the dataset and add more models in the future. 
  
\vspace{5 mm}
\section{Acknowledgments}
The authors thank IKEA Vaughan for granting permission to scan furniture items in their store. Special thanks are also due to Alexandru Andros, Gowri Nimal, Neshant Nimal, Nimal Sivapragasam, Prins Sivapragasam, Mano Sachi, Mithula Sachi, Sharmila Sachi, Sagnik Som, and Sachithananthan Subramanium for their generosity in permitting object scanning in their homes. 

\bibliographystyle{IEEEtran}
\bibliography{main}

\end{document}